\documentclass[runningheads]{llncs}

\RequirePackage{silence}
\usepackage[T1]{fontenc}
\usepackage{comment}
\usepackage{amsmath}
\usepackage{amssymb}
\usepackage{tabularx}
\usepackage{multirow}
\usepackage{booktabs}
\usepackage{graphicx}
\usepackage[table]{xcolor}
\usepackage{url}
\usepackage{hyperref}

\definecolor{tabgray}{gray}{0.94}
\newcommand{\myrowcolour}{\rowcolor{tabgray}}

\hypersetup{
  hidelinks,
  pdfauthor={Ondrej Valach, Vaclav Divis, Ivan Gruber},
  pdftitle={From Wrecks to Wisdom: Recovering Crash Mechanics from Real-World Multi-View Photos}
}

\newif\ifreview
\reviewfalse

\ifreview
    \usepackage{lineno}
    \renewcommand\thelinenumber{\color[rgb]{0.2,0.5,0.8}\normalfont\sffamily\scriptsize\arabic{linenumber}\color[rgb]{0,0,0}}
    
    \linenumbers
\fi

\begin{document}

%%%%%%%%%%%%%%%%%%%%% Add submission id, track, and title. %%%%%%%%%%%%%%%%%%%%%

% TODO: Insert the CMT submission number after registering the paper.
\def\SubNumber{44}

% TODO: Uncomment/change the track if you submit outside the regular main track.
\def\GCPRTrack{Main Track}

\title{From Wrecks to Wisdom: Recovering Crash Mechanics from Real-World Multi-View Photos}

\ifreview
    % ANONYMOUS SUBMISSION FOR REVIEW
    % DO NOT MODIFY these for the draft version used for review.
    \titlerunning{GCPR 2026 Submission \SubNumber{}. CONFIDENTIAL REVIEW COPY.}
    \authorrunning{GCPR 2026 Submission \SubNumber{}. CONFIDENTIAL REVIEW COPY.}
    \author{GCPR 2026 - \GCPRTrack{}}
    \institute{Paper ID \SubNumber}
\else
    % CAMERA READY SUBMISSION
    \titlerunning{From Wrecks to Wisdom}
    \author{
        Ond\v{r}ej Valach\inst{1}\orcidID{0009-0000-7629-0516}
        \thanks{Corresponding Author}
        \and
        V\'aclav Divi\v{s}\inst{1}\orcidID{0000-0001-9935-7824}
        \and
        Ivan Gruber\inst{1}\orcidID{0000-0003-2333-433X}
    }
    \authorrunning{O. Valach et al.}
    \institute{
        University of West Bohemia, Faculty of Applied Sciences,\\
        Department of Cybernetics and New Technologies for the Information Society\\
        \email{valacho@fav.zcu.cz, vincie@kky.zcu.cz, grubiv@ntis.zcu.cz}
    }
\fi

\maketitle

\begin{abstract}
Estimating accident mechanics from real-world crashes is important for vehicle-safety analysis, injury modeling, and crash-severity prediction. It can also support operational workflows such as insurance claim triage. In standard crash records, key metadata such as impact configuration, principal direction of force, and change in velocity ($\Delta V$) may be missing, delayed, or corrupted. By contrast, post-crash photographs are widely available and contain rich visual evidence of deformation. This raises the following question: how much crash mechanics information can be recovered directly from vehicle photos when structured signals are absent? To study this, we formulate crash understanding as supervised prediction from per-case multi-view photo sets. Targets include six Collision Deformation Classification (CDC) descriptors and the longitudinal/lateral components of reconstructed $\Delta V$. Each photo is encoded by a shared visual backbone, and the resulting view-level features are fused into a single case-level representation from which target-specific heads predict crash descriptors. Using 15.2k training cases from the Crash Investigation Sampling System ($\approx$1.5M photos before filtering), 1.15k validation cases, and 1.15k test cases, we define an evaluation protocol for vision-based crash descriptor estimation from incomplete multi-view evidence. Post-crash imagery alone provides usable signal for several non-trivial crash-mechanics descriptors, while weakly observable and long-tailed targets remain challenging. Within the compared training regimes, the selected joint-training recipe improves several context-dependent targets under this protocol, reducing mean absolute angular error for principal direction of force from $20.1^\circ$ to $14.05^\circ$ and lowering longitudinal $\Delta V$ MAE from $8.04$ to $7.45$~km/h. Our work provides a reference point for future multimodal fusion with structured crash metadata.

\keywords{crash severity prediction \and crash mechanics \and post-crash imagery \and multi-view learning \and computer vision \and deep learning \and traffic safety \and Collision Deformation Classification \and $\Delta V$ estimation \and multi-target prediction}
\end{abstract}

\section{Introduction}\label{intro}
    Traffic accidents remain a major public health and safety challenge, causing high rates of fatalities and injuries. Improving the analysis and prediction of crash outcomes is crucial for enhancing road safety, reducing casualties, and supporting more reliable crash reconstruction and interpretation. Active and passive safety systems, such as adaptive airbags, electronic stability control, and automated collision-avoidance systems, have improved substantially over the last decades. However, assessing how these systems perform in real-world crashes still requires accurate reconstruction of the underlying collision mechanics, information that is often absent or incomplete in standard crash records. Moreover, the parameters that describe these mechanics are often not directly measurable in real-world crashes, and many lower severity collisions go unreported (e.g., because they do not require a vehicle tow)~\cite{morando2025method}.
    
    \begin{figure}[t]
        \centering
        \includegraphics[width=0.9\linewidth]{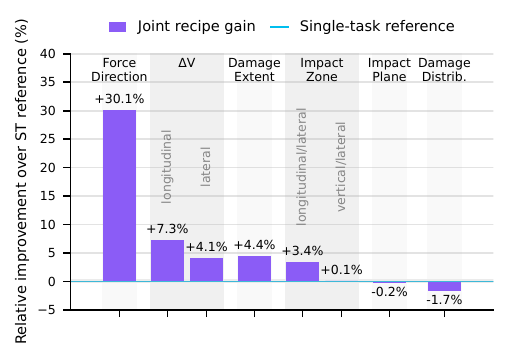}
        \caption{Comparison of the selected joint-training and single-task reference pipelines. Performance is shown as relative change (\%) with respect to the corresponding selected single-task result for each target.}
        \label{teaser}
    \end{figure}
    
    Beyond accident mechanics analysis itself, these crash descriptors are important inputs to downstream risk-modeling and triage pipelines. Unfortunately, they are typically available only after Event Data Recorder (EDR) extraction or crash reconstruction. By contrast, photographs are often among the earliest and most consistently available information channels. This motivates our study of image-based estimation from post-crash photo sets. In this work, we investigate how much of the underlying kinematic information can be recovered from images when structured signals are missing or delayed, and whether such estimates can serve as inputs to injury and severity models that would otherwise rely on structured metadata.

    Furthermore, we define crash understanding as multi-target prediction from post-crash photo sets. We predict deformation descriptors derived from the Collision Deformation Classification (CDC) code~\cite{sae_j224_202205} (Table~\ref{target_summary}) together with directional $\Delta V$ components. These targets capture complementary aspects of collision geometry and dynamics and are commonly used as informative inputs to downstream injury and severity models. To study this setting, we use a unified multi-view architecture built on a SwinV2 backbone~\cite{liu2022swin} with case-level fusion. We compare separate single-task training with a joint-training pipeline under a common input representation and evaluation protocol. As previewed in Figure~\ref{teaser}, the selected joint-training configuration improves several of the most context-dependent targets.

    In summary, this paper makes three contributions: (i) it introduces a CISS-based evaluation protocol for crash descriptor estimation from real-world, naturally incomplete multi-view post-crash photo sets; (ii) it defines a preprocessing and case-construction pipeline for mapping heterogeneous crash photographs into fixed canonical view slots with masked supervision for missing labels; and (iii) it reports single-task and joint-training reference results for CDC-derived crash descriptors and reconstructed $\Delta V$ components. Our goal is not to propose a new vision backbone, but to establish a controlled and reproducible reference point for image-based crash-mechanics estimation and future multimodal fusion with structured crash metadata.

\section{Related Work}
    Most prior crash-severity and injury prediction pipelines operate on structured crash records rather than images. Across this literature, kinematic variables such as change in velocity ($\Delta V$), principal direction of force, and peak acceleration are consistently treated as informative predictors alongside vehicle, occupant, roadway, and environmental factors~\cite{krafft2005influence,dean2023comparison,rifat2024explainable,niyogisubizo2021comparative,mostafa2025ai,gu2023injury}. In particular, prior studies identify $\Delta V$ and related crash-mechanics descriptors as important signals for injury estimation and severity modeling~\cite{krafft2005influence,dean2023comparison}. However, these quantities are usually assumed to be available from event data recorders, reconstruction pipelines, or investigator-coded records, rather than inferred directly from visual evidence.
    
    A smaller line of work addresses the missing-variable problem itself. Statistical reconstruction methods are used to correct reporting bias and infer unobserved crash quantities from structured records~\cite{morando2025method}. More broadly, most severity-prediction research still learns from tabular crash databases, police reports, and coded driver/vehicle/roadway variables, using both classical statistical models and modern machine-learning methods~\cite{niyogisubizo2021comparative,rifat2024explainable,mostafa2025ai,gu2023injury}. In these settings, crash-mechanics variables serve as inputs when available, not as prediction targets. Our focus is different: we study whether such descriptors can be recovered directly from post-crash photographs when structured signals are absent, delayed, or incomplete.
    
    An even more limited amount of work attempts to infer crash characteristics directly from post-crash images. Existing image-based methods typically focus on a narrow output space such as impact location or $\Delta V$, curated single-image settings, or supplement images with synthetic data or vehicle metadata~\cite{silver2022estimating,hasijahybrid}. The closest studies are Silver et al.~\cite{silver2022estimating}, who combine real and synthetic crash images, and Hasija et al.~\cite{hasijahybrid}, who use a curated frontal-only subset together with vehicle metadata. In contrast, we study real-world crash cases represented by naturally incomplete multi-view photo sets and predict multiple CDC-derived descriptors together with reconstructed $\Delta V$ components. Our goal is not to claim direct comparability with these settings, but to provide a reference point for image-based crash understanding at real-world scale.

\section{Dataset}\label{dataset}
    Datasets suitable for this task are limited. The problem requires crash records with post-crash photographs together with supervisory targets that describe collision mechanics. Nonetheless, we identified three repositories that provide this combination at scale: NHTSA's CISS/NASS-CDS in the United States, Germany's GIDAS, and India's RASSI~\cite{national2020overview,gidas_dataset_2023,rassi_dataset_2025}.

    We use NHTSA's Crash Investigation Sampling System (CISS) dataset, because it is openly available, large, standardized, and richly documented with post-crash imagery. Although GIDAS and RASSI also pair crash records with photographs, we excluded GIDAS due to access constraints and did not use RASSI because its road and traffic environment differs substantially from our target setting. Each retained CISS case includes structured forms and typically dozens of scene and vehicle photographs. From the available records, we construct a CISS-based evaluation setup of 17.5k crash cases ($\approx$1.5M photos before filtering).

    \begin{figure}[!ht]
        \centering
        \includegraphics[width=\linewidth]{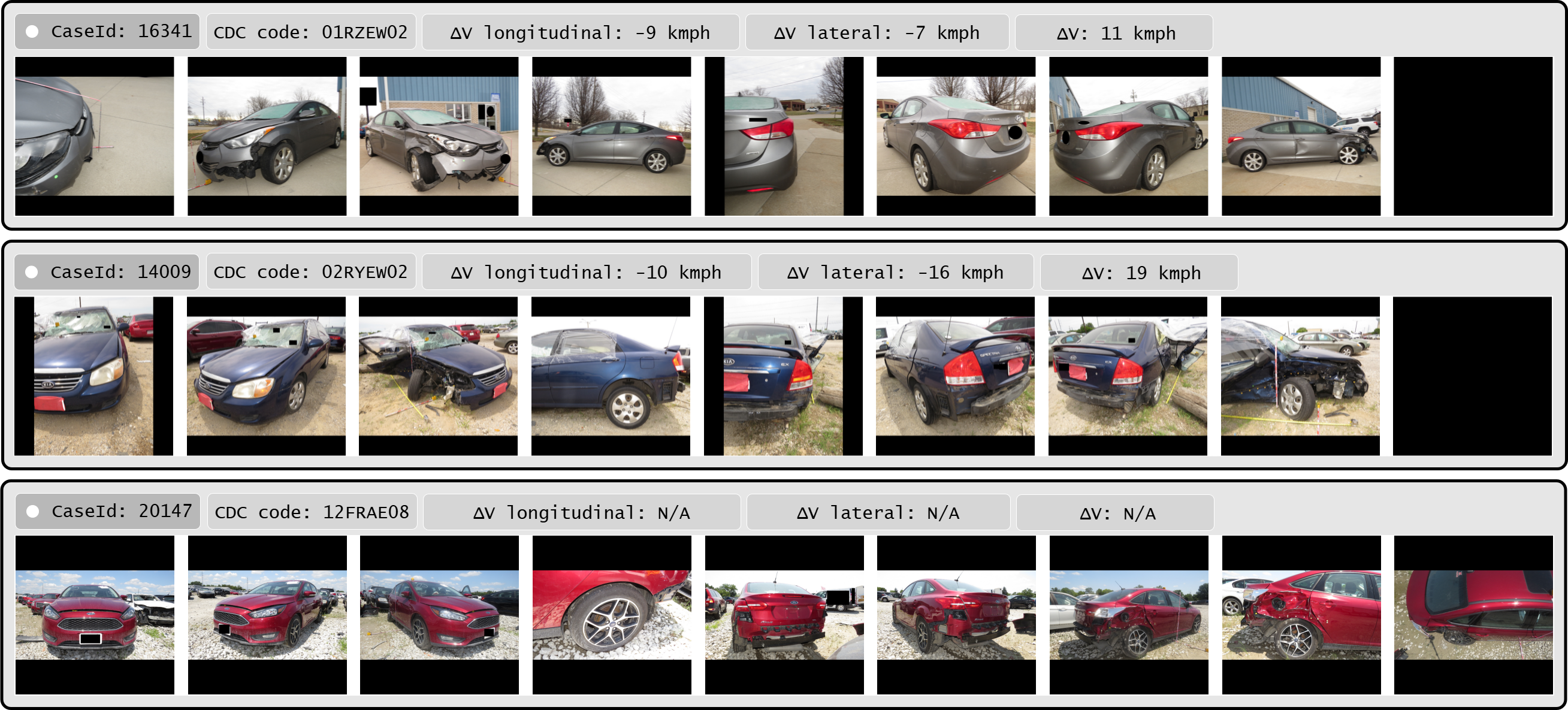}
        \caption{Representative crash cases in the fixed multi-view input format. Each case is shown as an ordered subset of available viewpoints together with its available crash descriptor annotations (CDC-derived targets and $\Delta V$ when available). The descriptor denoted as $\Delta V$ represents the resulting compound vector and is included only for reader interpretation.}
        \label{example_cases}
    \end{figure}

    For each case, the supervisory targets consist of an eight-character CDC code (investigator-coded) together with reconstructed WinSMASH $\Delta V$ components. WinSMASH is a crash reconstruction software, which works with scene evidence and vehicle damage information~\cite{sharma2007overview}. From the CDC code, we derive six targets: principal direction of force, deformation plane, longitudinal/lateral and vertical/lateral crush zones, damage distribution, and deformation extent. These targets and their evaluation metrics are summarized in Table~\ref{target_summary}.

    We train on $\Delta V_{\text{long}}$ and $\Delta V_{\text{lat}}$ as regression targets, while noting that these labels are reconstruction-derived estimates rather than direct measurements.

    The input to the model is an ordered multi-view photo set with up to nine canonical view slots: front, rear, right, left, front-right, front-left, rear-right, rear-left, and top. Cases are split at the crash-case level into disjoint train, validation, and test partitions to prevent leakage, with $N_\text{train}=15200$, $N_\text{val}=1150$, and $N_\text{test}=1150$. Because the target distributions are strongly long-tailed, reflecting the real-world crash statistics, we report both accuracy and tail-sensitive metrics such as macro-F1. The full label distributions are shown in the Electronic Supplementary Material (ESM), Figure~3.

    Figure~\ref{example_cases} illustrates representative cases in the fixed multi-view input format used throughout training and evaluation. Real-world crash photo sets also present several practical difficulties, including uninformative close-ups, missing viewpoints, and incomplete target annotations. We summarize these issues in Section~\ref{problem}, full preprocessing details are provided in the ESM, Section~2.

\section{Methods}\label{methods}
    Our goal is to predict crash descriptors from post-crash photo sets, so the core of the model is a shared visual encoder followed by case-level multi-view fusion. We use SwinV2 as the backbone for all post-crash views for three reasons. First, it provides hierarchical representations, preserving local cues in early stages while aggregating broader context, which is useful when the evidence is subtle, spatially localized, or low-contrast against the vehicle body (e.g., cracks, panel gaps, broken lamps). Second, its window-based self-attention captures longer-range interactions while remaining computationally feasible at the image resolution used in this study compared to global-attention vision transformers~\cite{dosovitskiy2020image}. Third, SwinV2 supports interpolation of pretrained positional information, allowing fine-tuning at a different input size while still benefiting from pretrained weights~\cite{liu2022swin}.

    Each crash case is represented by up to nine post-crash views, with the image order assigned using the dataset's viewpoint labels. When a canonical viewpoint is unavailable, the corresponding slot remains empty. This fixed ordering is important because it preserves viewpoint identity across cases and allows the fusion module to associate each feature vector with a consistent semantic position. We use metadata-based slots in the main experiments to keep crash descriptor prediction separate from possible viewpoint estimation errors. ESM Table~2 reports an automatic orientation-based slot assignment variant, which matches the metadata-based slots at the reported precision when averaged over three runs.
    
\begin{figure}[!ht]
        \centering
        \includegraphics[width=\linewidth]{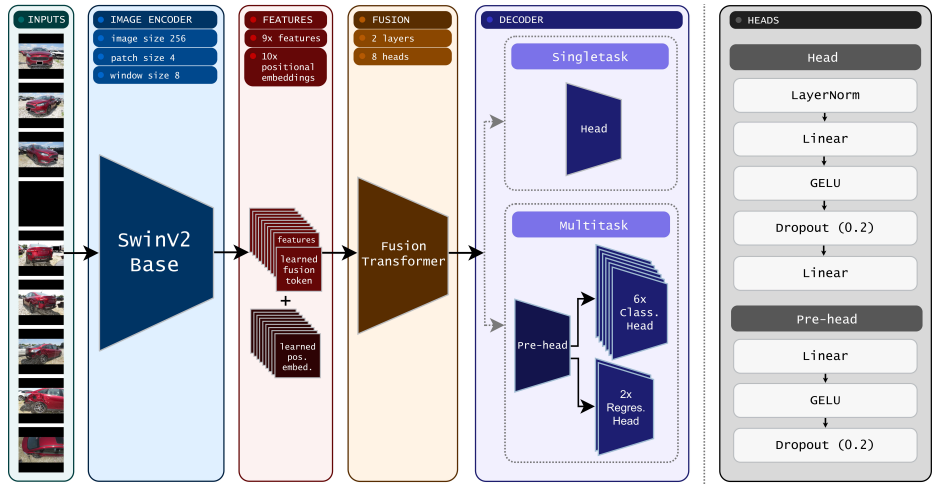}
        \caption{Multi-view crash descriptor architecture: a shared SwinV2 encodes nine ordered views, a lightweight fusion transformer aggregates them into a case embedding, and a decoder predicts CDC descriptors and $\Delta V$.}
        \label{arch}
    \end{figure}

    After the SwinV2 backbone, the architecture performs multi-view fusion, as shown in Figure~\ref{arch}. Each available view is processed independently by a shared SwinV2-base image encoder, producing one feature vector per view. These per-view features are treated as a token sequence and fused using a lightweight transformer~\cite{vaswani2017attention}: we prepend a learned fusion token and add learned positional embeddings so that the fusion module can exploit the consistent view ordering. After two transformer layers, the output is used as a single global representation of the crash case.
    
    On top of the fused embedding, we study two decoder configurations. In the single-task setting, the fused representation is passed to a single multilayer perceptron (MLP) prediction head that outputs one target at a time. In the multi-task setting, the backbone and fusion module remain shared, but the fused representation is first mapped through a shared pre-head into a common hidden space and then passed to multiple task-specific heads. Thus, the two settings share the same case representation, image backbone, and fusion module, while differing in output heads and in training strategy.
    
    \subsection{Preprocessing and Data Challenges}\label{problem}
    
    Real-world crash photo sets contain irrelevant close-ups, missing viewpoints, incomplete target annotations, and strongly imbalanced labels. We address these issues through vehicle and wheel filtering, fixed 9-slot view construction, view-drop augmentation, masked losses for missing labels, and imbalance-aware sampling. Full preprocessing, curation, and missing-label details are provided in the ESM, Section~2.
    
        \paragraph{Task Observability and Label Difficulty}\label{challenge}
        The targets differ substantially in visual observability. Impact plane and principal direction of force are often supported by the visible damage layout, whereas crush-zone descriptors and deformation extent may depend on subtle, occluded, or missing views. $\Delta V$ is the least directly observable target because it is a reconstructed kinematic quantity inferred only indirectly from post-crash appearance. Since labels are case-level and not localized to image regions, the model must identify useful evidence across heterogeneous multi-view inputs, which motivates joint training across related descriptors.

    \subsection{Method Setup}\label{method-setup}
    Preliminary ablations over backbones and view-fusion strategies favored independently encoded SwinV2 view tokens fused with a lightweight transformer. Details are reported in the ESM, Section~1. With this case representation fixed, the main study compares two practical training regimes: single-task models, where one descriptor is learned per model instance, and a joint-training model, where the shared trunk is trained with multiple task-specific heads.

\subsection{Descriptor Targets and Task Formulations}
\paragraph{Direction of Force}
The CDC principal direction of force (DoF) is modeled as a 12-class circular prediction problem. Since neighboring clock positions represent small angular deviations and opposite positions large ones, we replace one-hot labels with Gaussian-smoothed target distributions over circular distance and minimize KL divergence to the model log-probabilities. At evaluation, the predicted class is converted to an angle and scored by mean absolute angular error under minimal circular distance. We also tested $\sin(\theta),\cos(\theta)$ regression, but circular classification performed better in preliminary experiments; see the ESM, Section~1.

\paragraph{CDC categorical and zone descriptors}
For non-circular CDC descriptors, impact plane, damage distribution, longitudinal/lateral zone, and vertical/lateral zone, we use cross-entropy loss with the same imbalance-aware sampling scheme. We report accuracy and macro-F1, the latter reflecting performance on underrepresented classes. Deformation extent (01--09) is ordinal, so we train it as 9-way classification with label smoothing and report exact accuracy, Acc@$\pm1$ for predictions within one neighboring grade, and MAE in bin units to capture near-miss errors.

\paragraph{$\Delta V$ regression}
For $\Delta V$, we jointly regress the longitudinal and lateral components, $\Delta V_{\text{long}}$ and $\Delta V_{\text{lat}}$, using SmoothL1 (Huber) loss~\cite{girshick2015fast}. Because $\Delta V$ is missing for 38\% of cases, training and evaluation use only the remaining 62\% with valid labels. We report MAE and RMSE per component; the compound $\Delta V$ magnitude can be obtained as the Euclidean norm of the two predicted components.

\subsection{Single-task}
We use single-task training as a reference setting in which one descriptor is learned per model instance. All single-task models share the same case representation, image backbone, and fusion module, while the final prediction head is task-specific. This setup provides per-target reference results under a common input pipeline. Class imbalance is handled through task-specific sampling, and samples without the required target are skipped for that task. Full sampling and optimization details are provided in the ESM, Section~3.

\subsection{Multi-task}\label{multitask}
Several labels describe overlapping aspects of the same physical event, so joint training is a natural practical baseline: a single shared representation can support all descriptor heads and may encourage the model to encode global crash-configuration cues. We therefore compare independent single-task reference models with a joint-training configuration in which the SwinV2 backbone and fusion module are shared across targets. This comparison is intended to evaluate two practical training regimes under the same multi-view input protocol, not to isolate multi-task supervision as the only changing factor.

In the joint-training configuration, the fused case embedding is first passed through a shared pre-head MLP and then routed to task-specific heads, as shown in Figure~\ref{arch}. Missing labels are handled through task-specific loss masking, so each sample contributes only to the heads with valid annotations. Class imbalance is handled with a shared task-aware sampling policy, and loss balancing is stabilized with GradNorm and Dynamic Weight Averaging during training.

Because this configuration differs from the single-task reference not only in joint supervision but also in initialization, loss balancing, sampling, and optimization schedule, the comparison should be interpreted as a comparison between selected single-task and joint-training setups under a common input representation, not as a perfectly matched ablation of multi-task supervision alone. Full training details and hyperparameters are provided in the ESM, Section~3 and Table~3.

\section{Experiments and Results}\label{results}
This section reports the main experimental results. All models were trained on a single GPU (NVIDIA A40/L40S, depending on the run). The target definitions are summarized in Table~\ref{target_summary}.

To contextualize class imbalance, Table~\ref{target_summary} reports the share of the most frequent class among valid labels (Top1\%). For categorical targets, this corresponds to the accuracy of an always-majority rule under the matched train/test label distributions. We therefore interpret accuracy together with macro-F1 where applicable and treat small differences between selected training regimes cautiously.

\begin{table}[ht]
\centering

\caption{
Target summary with missing-label rates (Miss\%) and share of the most frequent class among valid labels (Top1\%). Circ-Smooth-Cls: circular classification with Gaussian-smoothed targets and KL divergence~\cite{kullback1951information}. Smooth-Cls: classification with label smoothing. Acc: accuracy. AngErr: mean absolute angular error (degrees). MAE: mean absolute error. Acc@$\pm$1: accuracy within one neighboring extent bin of the ground-truth label.
}
\label{target_summary}
\scriptsize
\setlength{\tabcolsep}{3pt}
\renewcommand{\arraystretch}{1.12}
\resizebox{\linewidth}{!}{
\begin{tabular}{@{}l l c c c c@{}}
\toprule
\multicolumn{1}{c}{\textbf{Target}} &
\multicolumn{1}{c}{\textbf{Out}} &
\multicolumn{1}{c}{\textbf{Classes}} &
\multicolumn{1}{c}{\textbf{Miss\%}} &
\multicolumn{1}{c}{\textbf{Top1\%}} &
\multicolumn{1}{c}{\textbf{Metrics}} \\
\cmidrule(r{0.2em}){1-1}
\cmidrule(lr{0.2em}){2-2}
\cmidrule(lr{0.2em}){3-3}
\cmidrule(lr{0.2em}){4-4}
\cmidrule(lr{0.2em}){5-5}
\cmidrule(l{0.2em}){6-6}
Impact plane
  & Classification
  & 4
  & 0.0
  & 64.9
  & Acc \\
\myrowcolour
DoF$_{\text{clock}}$
  & Circ-Smooth-Cls
  & 12
  & 2.9
  & 46.1
  & AngErr \\
Long./lat. zone
  & Classification
  & 9
  & 0.0
  & 38.3
  & Acc \\
\myrowcolour
Vert./lat. zone
  & Classification
  & 7
  & 0.0
  & 89.8
  & Acc \\
Damage distribution
  & Classification
  & 8
  & 0.0
  & 75.4
  & Acc \\
\myrowcolour
Deformation extent
  & Smooth-Cls
  & 9
  & 14.7
  & 43.6
  & Acc@$\pm$1 \\
$\Delta V_{\text{long}}$ / $\Delta V_{\text{lat}}$
  & Regression
  & -
  & 38.1
  & \textit{--}
  & MAE \\
\bottomrule
\end{tabular}
}
\end{table}

Unless stated otherwise, test results are reported at validation-selected checkpoints: best per head for single-task training and best joint objective for joint training. Because the reported single-task and joint-training models use different selected training configurations, summarized in the ESM, Table~3, the comparison should be read as an empirical comparison between two practical training regimes rather than as an isolated ablation of multi-task supervision. Accordingly, gains from the joint model reflect the selected joint-training pipeline as a whole, including shared supervision, initialization, sampling, and loss balancing.

\subsection{Single-task}\label{singletask}
We first report single-task baselines, where each descriptor is learned by an independent model using the same SwinV2 encoder and fusion transformer (Section~\ref{methods}). Test performance is reported at validation-selected checkpoints, and Table~\ref{multitask_compare} summarizes the results.

The single-task baselines show a clear split between visually well-supported descriptors and targets limited by weak evidence or label imbalance. Impact plane and DoF are learned well, indicating that the model can recover coarse impact geometry from post-crash views. The longitudinal/lateral zone also shows usable signal, suggesting that the model often localizes the main crush region along the vehicle length. In contrast, vertical/lateral zone and damage distribution are more strongly affected by long-tailed labels, with frequent classes dominating performance. Deformation extent is difficult to predict exactly, but near-miss agreement is substantially higher under Acc@$\pm1$, consistent with its graded severity interpretation. For $\Delta V$, the single-task model reaches MAE of 8.04 km/h for the longitudinal component and 5.31 km/h for the lateral component on the labeled subset, showing that visual evidence contains a measurable, though incomplete, signal for reconstructed kinematic quantities.

\subsection{Multi-task}\label{multitask_results}
We next report the joint model, which predicts all crash descriptors jointly using a shared SwinV2 encoder and fusion transformer with task-specific heads (Section~\ref{multitask}). The same evaluation protocol is used as for the single-task baselines, with the checkpoint selected by the validation joint objective. Table~\ref{multitask_compare} compares both regimes.

\begin{table}[ht]
\centering
\caption{
Comparison between selected single-task reference models and the selected joint-training recipe. $\Delta$ denotes the absolute change from the selected single-task model to the selected joint-training model, and Rel. $\Delta$ denotes the corresponding relative change with respect to the single-task result. For AngErr and MAE, lower is better. For Acc/$F1$, higher is better. The comparison should be interpreted as an empirical comparison between selected training regimes, not as an isolated ablation of multi-task supervision.
}
\label{multitask_compare}
\scriptsize
\setlength{\tabcolsep}{1.8pt}
\renewcommand{\arraystretch}{1.12}
\begin{tabular}{@{}l l c c c c@{}}
\toprule
\multicolumn{1}{c}{\textbf{Descriptor}} &
\multicolumn{1}{c}{\textbf{Metric}} &
\multicolumn{1}{c}{\textbf{Single-task}} &
\multicolumn{1}{c}{\textbf{Multi-task}} &
\multicolumn{1}{c}{$\boldsymbol{\Delta}$ \textbf{(ST$\rightarrow$MT)}} &
\multicolumn{1}{c}{\textbf{Rel.} $\boldsymbol{\Delta}$} \\
\cmidrule(r{0.2em}){1-1}
\cmidrule(lr{0.2em}){2-2}
\cmidrule(lr{0.2em}){3-3}
\cmidrule(lr{0.2em}){4-4}
\cmidrule(lr{0.2em}){5-5}
\cmidrule(l{0.2em}){6-6}

\multirow{2}{*}{Impact plane}
  & Acc & 0.924 & 0.922 & -0.002 & -0.2\% \\
  & $F1_{\mathrm{macro}}$ & 0.714 & 0.720 & +0.006 & +0.8\% \\

\myrowcolour
  & AngErr ($^\circ$) & 20.10 & 14.05 & -6.05 & -30.1\% \\
\myrowcolour
\multirow{-2}{*}{DoF$_{\mathrm{clock}}$}
  & Acc@$\pm$1 & 0.859 & 0.877 & +0.018 & +2.1\% \\

\multirow{2}{*}{Zone$_{\mathrm{Long./lat.}}$}
  & Acc & 0.582 & 0.602 & +0.020 & +3.4\% \\
  & $F1_{\mathrm{macro}}$ & 0.577 & 0.532 & -0.045 & -7.8\% \\

\myrowcolour
  & Acc & 0.901 & 0.902 & +0.001 & +0.1\% \\
\myrowcolour
\multirow{-2}{*}{Zone$_{\mathrm{Vert./lat.}}$}
  & $F1_{\mathrm{macro}}$ & 0.147 & 0.197 & +0.050 & +34.0\% \\

\multirow{2}{*}{Damage distribution}
  & Acc & 0.804 & 0.790 & -0.014 & -1.7\% \\
  & $F1_{\mathrm{macro}}$ & 0.290 & 0.293 & +0.003 & +1.0\% \\

\myrowcolour
  & Acc & 0.519 & 0.543 & +0.024 & +4.6\% \\
\myrowcolour
  & Acc@$\pm$1 & 0.823 & 0.859 & +0.036 & +4.4\% \\
\myrowcolour
\multirow{-3}{*}{Deformation extent}
  & MAE$_{\mathrm{bins}}$ & 0.939 & 0.818 & -0.121 & -12.9\% \\

$\Delta V_{\mathrm{long}}$
  & MAE & 8.04 & 7.45 & -0.59 & -7.3\% \\

$\Delta V_{\mathrm{lat}}$
  & MAE & 5.31 & 5.09 & -0.22 & -4.1\% \\
\bottomrule
\end{tabular}
\end{table}

The selected joint-training pipeline improves several context-dependent targets, with the largest gain on DoF angular error ($20.1^\circ \rightarrow 14.05^\circ$) and consistent improvements for deformation extent and both $\Delta V$ components. These results suggest that shared training can be useful in this setting, but the evidence should be interpreted at the pipeline level rather than as a clean causal estimate of multi-task supervision alone. The joint model differs from the single-task references in initialization, sampling, loss balancing, and optimization schedule, all of which may contribute to the observed changes.

The gains are also not uniform. Impact-plane accuracy changes only marginally, long./lat. zone macro-F1 decreases, and damage-distribution accuracy drops modestly. These failures are informative: they indicate that a single shared sampler and loss-balancing scheme may under-serve targets with different long-tail structure compared with task-specific single-task training. Thus, the joint model is best viewed as a compact reference pipeline with favorable aggregate behavior, not as uniformly superior across all descriptors. Detailed cross-paper reference points are provided in the ESM, Section~4.

\section{Conclusion}
We presented a CISS-based evaluation protocol for crash descriptor estimation from real-world, naturally incomplete multi-view post-crash photo sets. Using a unified SwinV2 and transformer-fusion architecture, we showed that post-crash imagery contains usable signal for recovering several CDC-derived deformation descriptors together with reconstructed longitudinal and lateral $\Delta V$ components. Under the selected training configurations used in this study, the integrated joint-training pipeline improved several targets, most notably reducing DoF angular error by 30.1\% (from $20.1^\circ$ to $14.05^\circ$), while also improving deformation extent and both $\Delta V$ components. These results should be interpreted as practical reference performance for image-based crash-mechanics estimation, not as a replacement for reconstruction or EDR-based analysis. At the same time, some descriptors remained limited by weak observability and strong class imbalance, highlighting that image-based crash understanding is feasible but still incomplete in this real-world setting. Overall, these results support post-crash imagery as a practical complementary source of crash-mechanics information and provide a reference point for future multimodal models that combine images with structured crash metadata.

A natural next step is multimodal crash understanding that combines photo evidence with structured crash metadata commonly used in severity modeling, including vehicle, occupant, roadway, and environmental factors. Such structured fields could improve robustness, especially for weakly observable descriptors and for $\Delta V$, where labels are noisy and incomplete. Architecturally, this suggests a multimodal design that (i) encodes images into a fused case embedding, (ii) embeds structured fields via a lightweight MLP or transformer, and (iii) fuses both modalities jointly. Another important direction is the construction of curated subsets with improved class balance, especially for rare crash descriptors that are difficult to assess reliably in the current long-tailed setting.

{
    \small
    \bibliographystyle{splncs04}
    \bibliography{database}
}

\end{document}

% TODO: Insert the same CMT submission number used in main.tex.
\def\SubNumber{44}

% TODO: Keep this aligned with main.tex.
\def\GCPRTrack{Main Track}

\title{\texorpdfstring{From Wrecks to Wisdom: Recovering Crash Mechanics from Real-World Multi-View Photos\\
\large Supplementary Material}{From Wrecks to Wisdom: Recovering Crash Mechanics from Real-World Multi-View Photos - Supplementary Material}}

\ifreview
    % ANONYMOUS SUBMISSION FOR REVIEW
    % DO NOT MODIFY these for the draft version used for review.
    \titlerunning{GCPR 2026 Submission \SubNumber{}. CONFIDENTIAL REVIEW COPY.}
    \authorrunning{GCPR 2026 Submission \SubNumber{}. CONFIDENTIAL REVIEW COPY.}
    \author{GCPR 2026 - \GCPRTrack{}}
    \institute{Paper ID \SubNumber}
\else
    % CAMERA READY SUBMISSION
    \titlerunning{From Wrecks to Wisdom}
    \author{
        Ond\v{r}ej Valach\inst{1}\orcidID{0009-0000-7629-0516}
        \thanks{Corresponding Author}
        \and
        V\'aclav Divi\v{s}\inst{1}\orcidID{0000-0001-9935-7824}
        \and
        Ivan Gruber\inst{1}\orcidID{0000-0003-2333-433X}
    }
    \authorrunning{O. Valach et al.}
    \institute{
        University of West Bohemia, Faculty of Applied Sciences,\\
        Department of Cybernetics and New Technologies for the Information Society\\
        \email{valacho@fav.zcu.cz, vincie@kky.zcu.cz, grubiv@ntis.zcu.cz}
    }
\fi

\maketitle

\section{Preliminary Experiments on Direction of Force (CDC Clock)}\label{sec:preliminary}

We ran preliminary experiments on the CDC 12-bin direction-of-force (DoF, clock) descriptor because it provides one of the clearest visual signals in post-crash photo sets and therefore served as a practical proxy for selecting the backbone and multi-view aggregation design. These experiments are intended as model selection evidence for the input and fusion pipeline.

We compared several backbones and multi-view aggregation strategies, and evaluated three practical ways of converting a crash case into a photo-set input: (i) sampling a random subset of $K$ images (with replication or permutation), (ii) concatenating all images along the channel dimension, and (iii) mapping photos into fixed ordered view slots with padding for missing viewpoints.

Across these experiments, tested CNN-style backbones (EfficientNet-B4 and InternImage-T) did not make sustained progress: training curves oscillated and validation angular error plateaued early. A modest improvement was observed when InternImage-T consumed all views jointly via channel concatenation, but performance still saturated quickly.

Transformer backbones adapted better to the task. With SwinV2-base, mean pooling provided a small gain, concatenation of per-view tokens improved further, and a dedicated token-fusion transformer operating on fixed ordered views produced the most stable learning behavior and the lowest validation error (about $20^\circ$ AngErr). This motivated the final design used in the main paper.

\begin{table}[ht]
\centering

\caption{
DoF ablations. Eff-B4: EfficientNet-B4. SwinV2-B: SwinV2-base. Intern-T: InternImage-T. Rand $K$: random subset of available views. Rep: replication if fewer than $K$ views. Perm: random permutation. Fixed-9: fixed ordered 9-slot input. Pad: zero-padding for missing views and edge padding to preserve image aspect ratio. AngErr: mean absolute angular error (degrees) under circular distance.
}
\label{tab:ablations_supp}
\scriptsize
\setlength{\tabcolsep}{2.2pt}
\renewcommand{\arraystretch}{1.08}
\begin{tabular}{@{}l l l l l c@{}}
\toprule
\multicolumn{1}{c}{\textbf{ID}} &
\multicolumn{1}{c}{\textbf{Backbone}} &
\multicolumn{1}{c}{\textbf{Views}} &
\multicolumn{1}{c}{\textbf{Fuse}} &
\multicolumn{1}{c}{\textbf{Target/Loss}} &
\multicolumn{1}{c}{\textbf{AngErr} ($^\circ$) $\downarrow$} \\
\cmidrule(r{0.2em}){1-1}
\cmidrule(lr{0.2em}){2-2}
\cmidrule(lr{0.2em}){3-3}
\cmidrule(lr{0.2em}){4-4}
\cmidrule(lr{0.2em}){5-5}
\cmidrule(l{0.2em}){6-6}
P1 & Eff-B4   & Rand $K$ (rep)  & Concat.     & $\sin/\cos$ MSE   & $44.1^\circ$ \\
\myrowcolour
P2 & Intern-T & Rand $K$ (rep)  & Concat.      & Circ-smooth KL    & $44.6^\circ$ \\
P3 & Intern-T & Chan. conc.     & --          & Circ-smooth KL    & $37.4^\circ$ \\
\myrowcolour
P4 & SwinV2-B   & Rand $K$ (perm) & Mean pool   & $\sin/\cos$ MSE   & $32.9^\circ$ \\
P5 & SwinV2-B   & Rand $K$ (rep)  & Concat.      & $\sin/\cos$ MSE   & $27.3^\circ$ \\
\myrowcolour
P6 & SwinV2-B   & Fixed-9 (pad)   & Transf.fuse & Circ-smooth KL    & $21.8^\circ$ \\
\bottomrule
\end{tabular}\end{table}

\subsection{Per-Setup Description (P1-P6)}
Table~\ref{tab:ablations_supp} summarizes the configurations and we describe each setup referenced by its P-code.

\paragraph{P1: EfficientNet-B4 + random multi-view replication + concatenation + circular regression}
P1 uses an EfficientNet-B4 encoder shared across views. Each crash case is represented by a random subset of $K$ available images; when fewer than $K$ views are present, images are replicated to keep a fixed view count (Rand $K$, rep). View features are fused by simple concatenation (Concat.), and the DoF target is trained via circular regression by predicting $(\sin\theta,\cos\theta)$ with an MSE loss. In practice, this configuration did not learn reliably: training was unstable and validation error oscillated around a plateau.

\paragraph{P2: InternImage-T + random multi-view replication + concatenation + circular smoothed classification}
P2 evaluates InternImage-T with the same randomized view construction (Rand $K$, rep). Per-view features are concatenated (Concat.) and the DoF head is trained as 12-way circular classification using Gaussian-smoothed targets over circular distance with KL divergence (Circ-smooth KL). Similar to P1, this per-view processing + concat setup showed limited learning progress and quickly saturated.

\paragraph{P3: InternImage-T + image per channel concatenation (single forward pass) + circular smoothed classification}
P3 changes only the input representation: instead of encoding views independently, images are stacked along the channel dimension and fed into InternImage-T in a single pass, so no explicit fusion module is needed. This produced a measurable but modest improvement over P2. However, validation error still plateaued relatively early, suggesting that the model benefited from joint low-level processing but remained limited in multi-view reasoning.

\paragraph{P4: SwinV2-base + random view permutation + mean pooling + circular regression}
P4 uses a SwinV2-base backbone, which processes each view as a token sequence. Views are sampled as a random subset with permutation (Rand $K$, perm), and fused by mean pooling over embeddings. The head predicts $(\sin\theta,\cos\theta)$ with an MSE loss. This configuration learned more consistently than CNN-style baselines, but improvements remained modest.

\paragraph{P5: SwinV2-base + random multi-view replication + concatenation + circular regression}
P5 keeps the SwinV2-base backbone but uses random-$K$ sampling with replication when fewer than $K$ views are available, and fuses the resulting view features by concatenation rather than mean pooling. This configuration achieved lower angular error than P4, suggesting that preserving per-view identity in the fused representation was beneficial, although the comparison also reflects the change in view construction.

\paragraph{P6: SwinV2-base + fixed ordered 9-view input + transformer fusion + circular smoothed classification}
P6 aligns with the final design direction: images are mapped into a fixed ordered 9-slot representation (Fixed-9), and missing canonical viewpoints are filled by padding. Per-view tokens are fused by a dedicated transformer-based fusion module (Transf.fuse), and the DoF head uses circularly smoothed classification with KL divergence (Circ-smooth KL). This setup produced the most stable learning behavior and reached the lowest validation error in these experiments ($21.8^\circ$ AngErr), motivating its use in the main model. Note that these numbers reflect results from the preliminary ablation study training runs, therefore the $21.8^\circ$ is not equal to our best single-task DoF result.

These ablations represent a compact subset of the exploratory experiments conducted during model development and are included because they capture the main design decisions that shaped the final model. Additional runs were performed, but they did not provide sufficiently distinct insights to justify separate inclusion here.

\section{Dataset Curation, Preprocessing, and Evaluation Details}\label{sec:preproc-details}
This section provides supplementary implementation details for preprocessing and dataset curation, example filtering outcomes, case construction, missing-label handling, view-drop augmentation, and checkpoint selection. These details support the main experimental protocol but are not essential for understanding the core architecture or main results.

\subsection{Crash Case Acquisition and Curation Pipeline}\label{sec:data-pipeline}

\begin{figure}[t!]
    \centering
    \includegraphics[width=\linewidth]{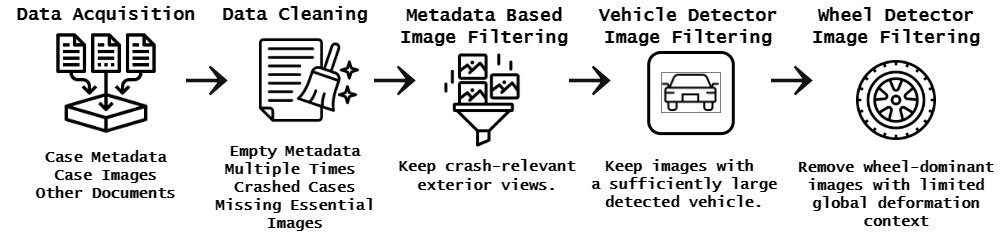}
    \caption{Overview of the crash case acquisition and curation pipeline. Starting from publicly available NHTSA CISS cases, we applied data cleaning, metadata-based image filtering, YOLOv8s vehicle filtering, and a dedicated RetinaNet-based wheel-detector filtering stage to obtain the final curated crash-image sets used in this study.}
    \label{fig:data_pipeline}
\end{figure}

Figure~\ref{fig:data_pipeline} summarizes the acquisition and curation pipeline used to construct the benchmark. We first collected publicly available crash investigation cases from the NHTSA CISS database, including case metadata, case images, and associated documents.

We then applied a data-cleaning stage to remove unusable cases, including entries with empty metadata, repeated or multiply crashed cases, and cases missing essential images. After this initial cleanup, we performed metadata-based image filtering to retain crash-relevant exterior vehicle views and discard visually uninformative categories.

The remaining candidate images were passed through a YOLOv8s vehicle-filtering stage, where we kept only images containing a sufficiently large detected vehicle. Finally, a dedicated RetinaNet-based wheel detector removed wheel-dominant close-up images that provide limited information about global crash deformation context. The resulting curated image sets were used for the downstream case construction and experiments reported in this work.

\subsection{Preprocessing and Data Challenges}\label{sec:preprocessing-challenges}

Given a case with post-crash images and annotations, we address four practical issues before training: irrelevant close-ups, missing viewpoints, missing target values, and strong class imbalance.

\begin{figure}[h]
    \centering
    \includegraphics[width=\linewidth]{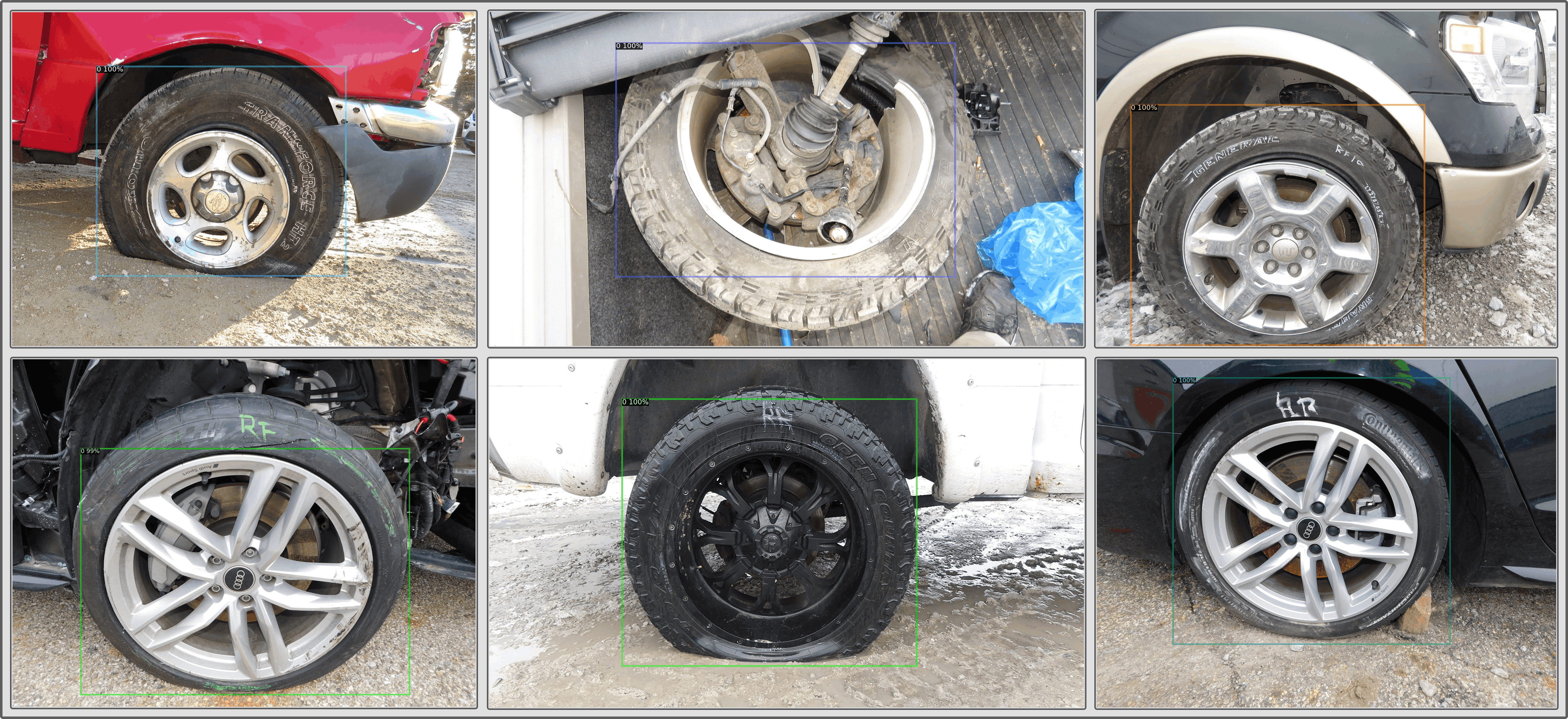}
    \caption{Representative examples removed by the wheel-filtering step. The discarded images are dominated by close-up wheel views and contain limited information about global crash deformation.}
    \label{fig:wheel_filter_examples}
\end{figure}

CISS vehicle photos sometimes include close-up images that carry little information about case-level deformation. A frequent example is a wheel close-up that is technically associated with a vehicle viewpoint but contributes little to global crash reasoning. To reduce this noise, we train a dedicated wheel detector model based on the RetinaNet~\cite{lin2017focal} architecture on the CAWDEC~\cite{novozamsky_cawdec} dataset and apply it as a preprocessing filter to remove images dominated by wheels. Representative removed examples are shown in Figure~\ref{fig:wheel_filter_examples}.

Most cases are missing one or more canonical viewpoints, so the corresponding slots remain empty. During training, we additionally apply random view-drop augmentation to improve robustness to incomplete evidence. At inference, the model receives only the available post-crash images.

Some cases also lack target values, most commonly $\Delta V$. In single-task training, such samples are skipped for the relevant head. In multi-task training, losses are masked per head so that a sample contributes only to the targets with valid annotations, preserving partially annotated cases. Finally, because the CDC-derived targets are strongly imbalanced, we handle long-tail structure during sampling and report macro-F1 in addition to accuracy.

\subsection{Wheel-Detector Filtering Details}\label{sec:wheel-filter}

To reduce noise from close-up images that do not capture the global deformation pattern, we trained a dedicated wheel detector on the CAWDEC dataset~\cite{novozamsky_cawdec}. The detector was used only as a preprocessing filter: images dominated by wheels were removed before case-level view construction. This was necessary because such close-ups are not identifiable from the available CISS metadata alone, even though they carry little information about crash-level deformation.

In the preprocessing pipeline, the detector was applied to candidate vehicle images prior to slot assignment. Images identified as wheel-dominated were excluded from further processing, while ordinary side views that still contained broader crash evidence were retained. Overall, this step reduced visually uninformative close-ups while preserving images relevant to case-level deformation reasoning.

\subsection{9-Slot Construction}\label{sec:view-slots}

Each crash case was mapped into a fixed 9-slot representation corresponding to the canonical viewpoints used throughout the study: front, rear, right, left, front-right, front-left, rear-right, rear-left, and top. When a viewpoint was unavailable, the corresponding slot was left empty and filled by padding. This representation preserves viewpoint identity across cases and allows the fusion module to associate tokens with consistent semantic positions.

If multiple images were available for the same canonical slot, the preprocessing pipeline selected one representative image at random. The same ordered slot representation was used during both training and evaluation. To make this step reproducible, the random choice is fixed by the preprocessing seed before model training, so validation and test cases use deterministic slot assignments. Since the slot assignment uses viewpoint metadata, this setup should be understood as image-based prediction with view organization supported by metadata, not as fully unconstrained image-set learning. For datasets without viewpoint metadata, we also evaluate automatic slot assignment using a vehicle orientation estimator trained from the Car Full View Dataset~\cite{catruna2023car}. Table~\ref{tab:slot_assignment_ablation} compares this automatic variant with the metadata-based slot assignment used in the main benchmark.

\begin{table}[t]
\centering
\caption{
Effect of view-slot construction on DoF prediction. Metadata slots use CISS viewpoint labels and are used in the main experiments to isolate crash-descriptor learning from viewpoint-estimation noise. Auto-orientation slots use an image-based vehicle-orientation estimator trained from the Car Full View Dataset~\cite{catruna2023car} to assign views to the fixed 9-slot representation. Results are averaged over three runs using the same DoF model and evaluation protocol.
}
\label{tab:slot_assignment_ablation}
\scriptsize
\setlength{\tabcolsep}{3pt}
\renewcommand{\arraystretch}{1.08}
\begin{tabular}{@{}l l c@{}}
\toprule
\textbf{Slot construction} & \textbf{Input organization} & \textbf{AngErr} ($^\circ$) $\downarrow$ \\
\midrule
Random / unordered & Random available views & 25.18 \\
Metadata slots & Fixed 9-slot, CISS metadata & 20.10 \\
Auto-orientation slots & Fixed 9-slot, predicted orientation & 20.10 \\
\bottomrule
\end{tabular}
\end{table}

Both fixed-slot variants substantially reduce DoF angular error compared with unordered random view construction. Averaged over three runs, the auto-orientation result matches the metadata-slot result at the reported precision, suggesting that the gain comes mainly from organizing views into consistent canonical slots. For DoF prediction, this indicates that predicted vehicle orientation can replace viewpoint metadata for this part of the pipeline.

\subsection{View-Drop Augmentation and Missing Labels}\label{sec:view-drop}

During training, we randomly dropped between 0 and 6 available views per case to improve robustness to incomplete evidence. This augmentation exposes the model to varying levels of view availability during training. At inference, no views are synthetically dropped and the model receives only the available post-crash images.

Missing target labels were handled differently in the two regimes. In single-task training, samples without the required target were skipped for that task. In multi-task training, losses were masked per head so that a sample contributed only to the targets with valid annotations.

\subsection{Target Coverage and Label Imbalance}\label{sec:target-distributions}

\begin{figure}[!t]
    \centering
    \includegraphics[width=\linewidth]{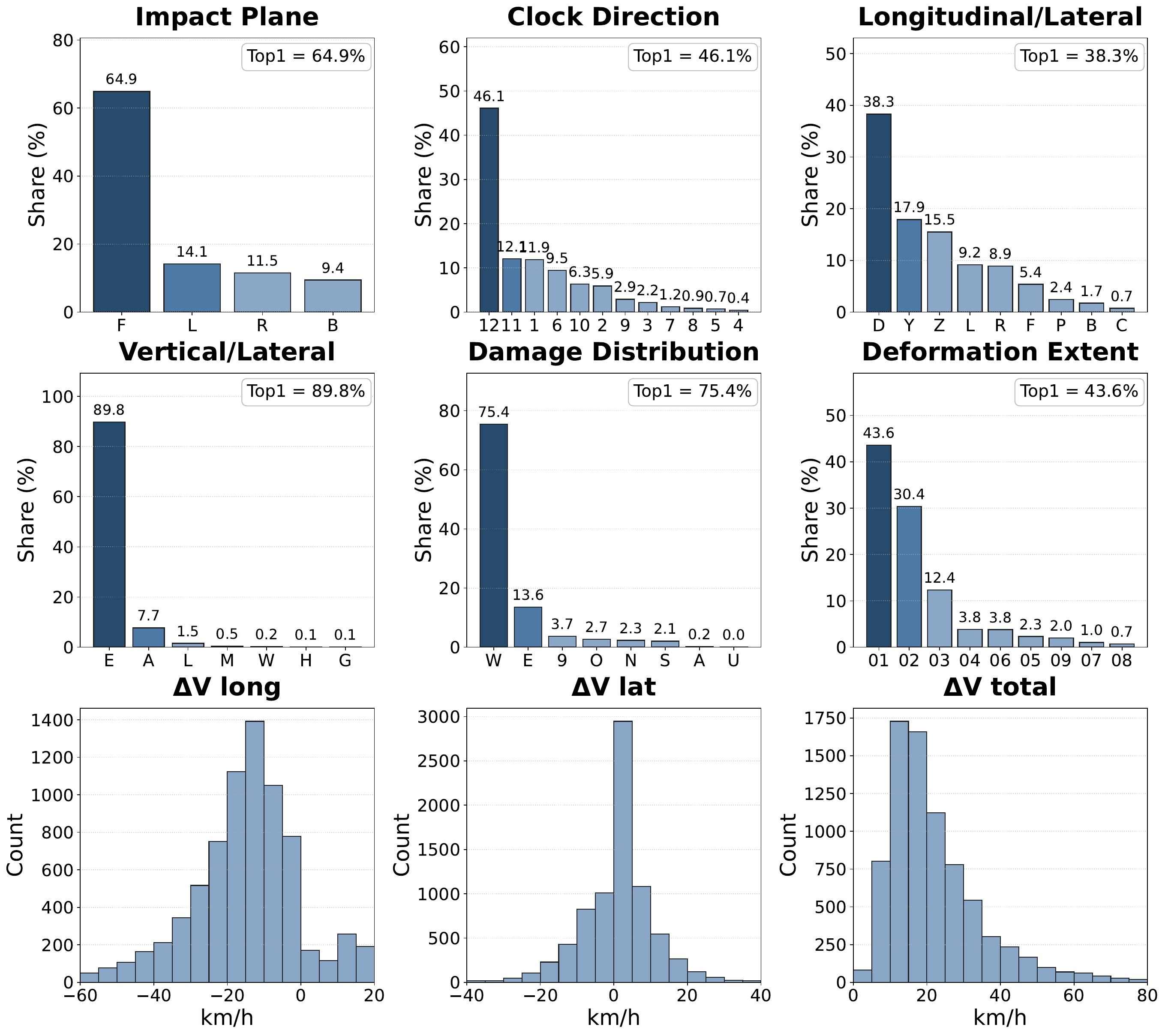}
    \caption{Label distributions in our CISS-based dataset. We show the class frequencies of all CDC-derived targets together with the distributions of $\Delta V_{\text{long}}$ and $\Delta V_{\text{lat}}$ (WinSMASH reconstruction). The compound $\Delta V$ vector is included for reader interpretation. The categorical targets show strong long-tail behavior with dominant frontal/central configurations.}
    \label{fig:output-distributions}
\end{figure}

Figure~\ref{fig:output-distributions} visualizes the empirical distributions of all targets used in this study, including CDC-derived categorical descriptors and reconstructed WinSMASH $\Delta V$ components. The categorical targets exhibit pronounced long-tail behavior, with frontal and centrally located configurations occurring much more frequently than rarer side, corner, and uncommon deformation patterns.

Qualitative failure cases generally fall into four categories: missing or occluded views of the primary damaged region, visually subtle deformation despite non-negligible reconstructed $\Delta V$, rare CDC classes with few training examples, and cases where close-up or partial vehicle images dominate the available photo set. These patterns are consistent with the quantitative results: coarse impact geometry is often visually recoverable, whereas fine-grained crush zones, damage distribution, deformation extent, and $\Delta V$ can depend on evidence that is missing, weakly visible, or only indirectly encoded in post-crash appearance.

\section{Training and Optimization Details}\label{sec:training-details}

This section provides the training and sampling details used for the single-task and multi-task configurations reported in the main paper. These details are important for reproducibility, but they are not central to the main methodological narrative.

\subsection{Single-Task Training Details}\label{sec:single-task-training-details}

In the single-task setting, one descriptor is learned per model instance. All single-task models use the same case representation, image backbone, and fusion module. Only the final head and task-specific training configuration differ. This setup provides per-target reference results under a common input pipeline, while allowing label-specific sampling choices for strongly imbalanced targets.

Most descriptor targets are categorical, so we formulate them as supervised classification and address long-tailed labels during sampling. To mitigate class imbalance, we use inverse-frequency sampling together with a stratified window scheme aligned with gradient accumulation. Specifically, samples are weighted by inverse label frequency, optionally raised to a power. Rare labels are defined by a high-quantile threshold, usually 0.90, on these weights, and each accumulation window, i.e., the set of micro-batches forming one optimizer update, is constrained to contain a minimum number of rare samples. The exact sampling setup is tuned separately for each label.

A practical advantage of the single-task setting is that the sampling policy can be tailored to each descriptor, allowing batches to approach a more balanced label distribution without additional dataset curation. In contrast, a single batch in multi-task training must serve several differently imbalanced targets at once, making long-tail mitigation more difficult.

Across all tasks, we used AdamW~\cite{loshchilov2017decoupled}, a Trapezoidal learning-rate scheduler~\cite{xing2018walk}, and an input resolution of $640\times640$ pixels.

\subsection{Multi-Task Training Details}\label{sec:multi-task-training-details}

In the joint-training configuration, the fused case embedding is first passed through a shared pre-head MLP and then routed to task-specific heads. The shared trunk is initialized from the single-task DoF$_{\text{clock}}$ pretrained model checkpoint. Newly added task-specific layers are initialized randomly, and missing labels are handled through task-specific loss masking.

Because this configuration differs from the single-task reference not only in joint supervision but also in initialization, loss balancing, sampling, and optimization schedule, the comparison should be interpreted as a comparison between selected single-task and joint-training setups under a common input representation, not as a perfectly matched ablation of multi-task supervision alone.

A major difficulty in the joint-training setting is that all heads must share a single sampling policy, even though the categorical targets are imbalanced in different ways. A standard inverse-frequency sampler is therefore not sufficient, because rarity is task-dependent: the same sample may be rare for one target and common for another. We therefore use task-aware weighted sampling. For each sample, we compute inverse-frequency weights for the categorical targets with available labels and take their maximum as the sampling weight. This increases the sampling probability of cases that are rare for at least one task. Sampling is then performed with replacement according to these weights.

While this does not guarantee balanced batches for every target, it improves exposure to underrepresented cases while maintaining a single shared training stream for all heads. From epoch 6 onward, we additionally enable GradNorm~\cite{chen2018gradnorm} and Dynamic Weight Averaging (DWA)~\cite{liu2019end} to further stabilize gradient balance across task heads. These mechanisms are part of the reported joint-training recipe and should not be separated from the empirical comparison to the single-task references.

\begin{table}[t]
\centering
\caption{Representative training configurations for the reported single-task and multi-task results. Both regimes use the same case representation, backbone, and fusion module, but differ in initialization, optimization, and loss balancing. Single-task hyperparameters were tuned per target, so that column is a compact summary rather than one identical configuration for all descriptors. Quant.-strat. rare/update: quantile-stratified update-level sampling with at least one rare sample per effective optimizer update. Task-aware WRS: WeightedRandomSampler using the maximum available inverse-frequency weight across categorical targets. Init. strategy: transfer initializes from a related single-task checkpoint, e.g., impact plane for vertical/lateral finetuning; clock-pretrain + random initializes the shared trunk from DoF$_{\text{clock}}$ pretraining and the remaining components randomly.}
\label{tab:train_setup}
\scriptsize
\setlength{\tabcolsep}{3.2pt}
\renewcommand{\arraystretch}{1.10}
\begin{tabular}{@{}lcc@{}}
\toprule
& \multicolumn{1}{c}{\textbf{Single-task}} & \multicolumn{1}{c}{\textbf{Multi-task}} \\
\cmidrule(r{0.2em}){1-1}
\cmidrule(lr{0.2em}){2-2}
\cmidrule(l{0.2em}){3-3}
Epochs & 50 & 50 \\
\myrowcolour
Learning rate & $1\!\times\!10^{-5}$ & $2/6/30\!\times\!10^{-6}$ \\
LR split & -- & backbone / fusion+pre-head / task heads \\
\myrowcolour
Warmup / anneal & 3 / 7 & 5 / 20 \\
Batch / accum. & 4 / -- & 1 / 4 \\
\myrowcolour
Effective batch & 4 & 4 \\
Views / mask & 9 / $k \in \{0,\ldots,6\}$ & 9 / $k \in \{0,\ldots,6\}$ \\
\myrowcolour
Sampler & quant.-strat. rare/update & task-aware WRS \\
Loss & CE / KL / SmoothL1 & CE + KL($\sigma{=}0.85$) + SmoothL1 \\
\myrowcolour
DWA & -- & yes \\
GradNorm & -- & from epoch 6 \\
\myrowcolour
Init. strategy & random / transfer & clock-pretrain + random \\
\bottomrule
\end{tabular}
\end{table}

\section{Reference Points to Prior Image-Based Work}\label{sec:refpoints}
Most crash-severity and injury pipelines rely on structured metadata (EDR, reconstruction outputs, or curated case forms). In contrast, we study an image-based setting at real-world scale, with naturally incomplete multi-view photo sets. The closest prior image-based crash studies differ substantially in target definitions, crash-mode coverage, curation, and input modality. In particular, Silver et al.~\cite{silver2022estimating} combine curated real-world photographs with synthetic images generated in the Rigs of Rods simulator~\cite{ohlidal2020rigs}, while Hasija et al.~\cite{hasijahybrid} use curated real-world cases together with vehicle metadata from NHTSA sources. We therefore report their numbers as reference points rather than as direct baselines.
\paragraph{Impact plane prediction.}
Table~\ref{impact_reference} reports comparisons for impact plane prediction, comparing prior single-image impact location with our CDC impact plane estimation from multi-view photo sets.

\begin{table}[ht]
\centering
\caption{
Impact plane image prediction. Silver et al.~\cite{silver2022estimating} evaluate in a curated single-image setting where the background is masked to retain only the vehicle, inputs are limited to front or rear views, and the label space is analogously reduced to front/back location. In contrast, our benchmark uses real-world crash cases represented as multi-view photo sets with natural viewpoint missingness and 4-plane coverage (front/right/left/back).
}
\label{impact_reference}
\scriptsize
\setlength{\tabcolsep}{2.0pt}
\renewcommand{\arraystretch}{1.08}
\begin{tabularx}{\columnwidth}{@{}Y Y Y c@{}}
\toprule
\textbf{Work} & \textbf{Input} & \textbf{Target} & \textbf{Acc} \\
\cmidrule(r{0.2em}){1-1}
\cmidrule(lr{0.2em}){2-2}
\cmidrule(lr{0.2em}){3-3}
\cmidrule(l{0.2em}){4-4}
Silver et al.~\cite{silver2022estimating} & 1 img & LOC (f/b) & 0.920 \\
\myrowcolour
Ours (ST) & $\leq 9$ images & CDC (f/r/l/b) & 0.924 \\
Ours (MT) & $\leq 9$ images & CDC (f/r/l/b) & 0.922 \\
\bottomrule
\end{tabularx}
\end{table}

These numbers are included only to contextualize task difficulty. Because the label spaces, curation protocols, and inputs differ, they should not be interpreted as evidence of performance parity.

\paragraph{$\Delta V$ estimation.}
Table~\ref{deltav_reference} reports reference point results for image-based $\Delta V$ prediction. Because prior work differs in data curation, crash-mode coverage, available metadata, and evaluation metrics (MAE/RMSE), these comparisons should not be interpreted as direct baselines. Instead, the table is intended to contextualize our real-world image-based setting and to show that post-crash imagery provides a usable signal that could support downstream injury, crash severity, or metadata prediction pipelines.

\begin{table}[ht]
\centering

\caption{
$\Delta V$ estimation results. Scope codes: R1 (Silver et al.~\cite{silver2022estimating}) uses a manually curated set (combined from simulation and real world) of small passenger vehicles for front/rear collisions ($<96$\,km/h), images are grayscale and manually cropped to the vehicle, reporting MAE for $\Delta V_{\text{long}}$. R2 (Hasija et al.~\cite{hasijahybrid}) uses a curated frontal-only subset and incorporates vehicle metadata (weight, body type, stiffness); we report their main RMSE for $\Delta V_{\text{long}}$ on the full test set (6.38 km/h). They additionally report 7.50 km/h on the EDR-verified subset. R3 (ours) evaluates on real-world CISS multi-view photo sets with natural missingness and image-based inputs, we predict component-wise $(\Delta V_{\text{long}},\Delta V_{\text{lat}})$ and report MAE and RMSE, with missing labels handled by masking and a 4-way impact plane.
}
\label{deltav_reference}
\scriptsize
\setlength{\tabcolsep}{1.4pt}
\renewcommand{\arraystretch}{1.08}
\begin{tabularx}{\columnwidth}{@{}>{\raggedright\arraybackslash}X p{0.17\columnwidth} p{0.08\columnwidth} >{\raggedright\arraybackslash}X >{\raggedright\arraybackslash}X r@{}}
\toprule
\multicolumn{1}{c}{\textbf{Work}} &
\multicolumn{1}{c}{\textbf{Input}} &
\multicolumn{1}{c}{\textbf{Scope}} &
\multicolumn{1}{c}{\textbf{Metric}} &
\multicolumn{1}{c}{\textbf{Target}} &
\multicolumn{1}{c}{\textbf{Score [km/h]}} \\
\cmidrule(r{0.2em}){1-1}
\cmidrule(lr{0.2em}){2-2}
\cmidrule(lr{0.2em}){3-3}
\cmidrule(lr{0.2em}){4-4}
\cmidrule(lr{0.2em}){5-5}
\cmidrule(l{0.2em}){6-6}
Silver et al.~\cite{silver2022estimating} & 1 image & R1 & MAE & $\Delta V_{\text{long}}$ & 4.19 \\
\myrowcolour
Hasija et al.~\cite{hasijahybrid} & 1 image, meta & R2 & RMSE & $\Delta V_{\text{long}}$ & 6.38 \\
& & & MAE & $(\Delta V_{\text{long}}, \Delta V_{\text{lat}})$ & 8.04 / 5.31 \\
\multirow[c]{-2}{*}{Ours (ST)} & \multirow[c]{-2}{*}{$\leq 9$ images} & \multirow[c]{-2}{*}{R3} & RMSE & $(\Delta V_{\text{long}}, \Delta V_{\text{lat}})$ & 11.96 / 8.20 \\
\myrowcolour
& & & MAE & $(\Delta V_{\text{long}}, \Delta V_{\text{lat}})$ & 7.45 / 5.09 \\
\myrowcolour
\multirow[c]{-2}{*}{Ours (MT)} & \multirow[c]{-2}{*}{$\leq 9$ images} & \multirow[c]{-2}{*}{R3} & RMSE & $(\Delta V_{\text{long}}, \Delta V_{\text{lat}})$ & 11.26 / 7.25 \\
\bottomrule
\end{tabularx}\end{table}

Taken together, these reference points help position our study within the body of prior image-based crash work. Because the target definitions, curation protocols, crash-mode coverage, and available inputs differ substantially across works, we use these comparisons only for context rather than as direct baselines. Within that constraint, our results provide a practical reference point for crash descriptor estimation from real-world, incomplete multi-view photo sets and motivate multimodal fusion as a natural next step, rather than establishing performance parity with prior curated or metadata-assisted settings.

{
  \small
  \bibliographystyle{splncs04}
  \bibliography{database}
}